# Deep Image Feature Learning with Fuzzy Rules

Xiang Ma, Liangzhe Chen, Zhaohong Deng, *Senior Member, IEEE*, Peng Xu, Qisheng Yan, Kup-Sze Choi, Shitong Wang

***Abstract*—Feature extraction methods are key to many image processing tasks. At present, the most popular method is to use a deep neural network to automatically extract features through end-to-end training instead of the traditional hand-crafted feature extraction. However, the training of deep neural network relies heavily on data quality and quantity, and the network is a black-box model that has poor interpretability. Human intelligence can be leveraged here to improve the deep neural network model, where the human decision process can be integrated in feature learning and object classification to enhance robustness and interpretability. In this paper, the method Deep Image Feature Learning with Fuzzy Rules (DIFL-FR) is proposed, where human decision process is embedded in feature extraction by combining fuzzy logic rule-based modeling with deep-stacked learning strategy. The proposed method has the following distinctive characteristics. First, since the method is based on fuzzy sets and fuzzy inference, it can extract more robust features from noisy data scenes. Second, the method progressively learns the image features through a layer-by-layer approach based on fuzzy rules, so that the feature learning process can be better explained by the rules generated. Third, the learning process of the method has a high efficiency since it is only based on forward propagation without back propagation and iterative learning. Finally, while the method is based on unsupervised learning, it can be easily extended to supervised and semi-supervised learning cases. The results of extensive experiments conducted on image datasets of different scales clearly show the effectiveness of the proposed method.**

## I. Introduction

IMAGE feature learning is a basic research topic in the field of computer vision and machine learning. It is the initial step in various tasks of computer vision, such as image classification, object detection and scene segmentation, followed by other techniques to achieve their respective goals. The design of an effective image feature learning method is nontrivial since the robustness of the learned features can be influenced by many factors, including occlusion, distortion and scaling.

At present, image feature extraction methods can be broadly divided into handcraft-based methods and learning-based methods (e.g., using machine learning or deep learning).

*Handcraft-based methods:* This category of methods extracts both global and local features. Examples of global features includes the region of interest and the background that provide global information. The most representative global feature extractors are Histogram of Oriented Gradient (HOG) [1], Local Binary Pattern (LBP) [2], color histograms [48], etc. The local features describe the internal information and details of the images. A series of methods based on the bag of words [3] approach have been proposed to extract local features, such as soft quantization [4, 5], locality-constrained linear coding [6] and spatial pyramid [7].

*Learning-based methods:* This category of methods automatically obtain features without using elaborately designed feature extractors. Compared with handcraft-based methods, learning-based methods can learn features directly from the image data and can better reveal the intrinsic information of the data. Learning-based methods are mainly based on matrix decomposition techniques and deep learning. For the former, matrix decomposition is used to find a mapping is to transform the high-dimensional image data into a low-dimensional space, from which the geometric structure information of the data is utilized. It is assumed that high-dimensional image data are embedded in a low-dimensional manifold in the high-dimensional space. The commonly used matrix decomposition methods include Vector Quantization (VQ) [8], QR decomposition [9], Singular Value Decomposition (SVD) [10], and Nonnegative Matrix Factorization (NMF) [11]. Turk et al. proposed the eigenface method [12], which applied Principal Component Analysis (PCA) [13] for face recognition. Ronald et al. proposed the Fisherface [14] model, which used Linear Discriminant Analysis (LDA) [15] to find the projection direction that maximizes the between-class scatter and minimizes the within-class scatter. The methods based on deep learning have received extensive attention in recent years. These methods extract abstract and effective high-level information by combining the low-level features to discover different feature representations of the data [16]. For example, Hinton et al. proposed the Deep Belief Network (DBN) [17], which is a generative model that This work was supported in part by the National key R & D plan under Grant (2022YFE0112400), the NSFC (62176105, 71961001), the Hong Kong Research Grants Council (PolyU 152006/19E) and CAAI-Huawei MindSpore Open Fund (CAAIXSJLJJ-2021-011A). (*Corresponding author: Zhaohong Deng; Xiang Ma and Liangzhe Chen have the equal contributions*).

X. Ma, L. Z. Chen, Z. H. Deng, P. Xu, S. Wang are with the School of Digital Media, Jiangnan University and Jiangsu Key Laboratory of Digital Design and Software Technology, Wuxi 214122, China (e-mail: 6171610013@stu.jiangnan.edu.cn; 2639885989@qq.com; dengzhaohong@jiangnan.edu.cn; 6171610015@stu.jiangnan.edu.cn; wxwangst@aliyun.com).

Q. Yan is with the School of Science, East China Institute of Technology, Nanchang 330013, China (e-mail: yanqs93@163.com).

K. S. Choi is with The Centre for Smart Health, the Hong Kong Polytechnic University (e-mail: thomasks.choi@polyu.edu.hk).

can extract high-level visual features of images. Convolutional Neural Networks (CNNs), with fewer network parameters and simplified training, have been proposed and extensively used in image processing [18]. The convolutional architectures of CNN are a key success factor for image feature extraction. A variety of techniques have been developed to learn the filter bank at each stage of a CNN. Regularized autoencoders or their variants [19] are among the typical approaches.

The handcraft- and learning-based methods have distinctive advantages in various image processing tasks. However, they all have their own defects. On the one hand, the handcraft-based methods are too sensitive to target occlusion, distortion and scaling. As image data are usually voluminous, high dimensional, unstructured, and exhibit uncertainty, it laborious, and difficult to design handcraft-methods [19]. On the other hand, learning-based methods such as the Deep Neural Networks (DNNs) requires a sound hypothesis space that needs to be driven by a large amount of data. With small data, deep neural networks are prone to overfitting, leading to poor performance and generalizability. In addition, deep back propagation [20] algorithms are usually required for training neural networks. For networks with a large number of layers, the vanishing gradient problem occurs, where long training time is needed for the model to reach convergence. Deep neural network is typically trained with the stochastic gradient descent (SGD) [21] method. The performance of the resulting model depends seriously on parameter tuning expertise and ad hoc tricks. A further issue of neural networks is the black-box nature of the models which lack interpretability.

To overcome the above drawbacks of the existing image feature methods, the Deep Image Feature Learning with Fuzzy Rules (DIFL-FR) is proposed. The advantages of DIFL-FR are three-fold. First, it is an uncertainty-system-based image feature learning method that is less sensitive to noise and occlusion in images. Second, it is also a rule-based system based on Takagi-Sugeno-Kang fuzzy system (TSK-FS) that is more transparent due to the improved interpretability of the model structure with fuzzy rules and the fuzzy inference mechanism. Specifically, TSK-FS [22–24] is taken as a shallow feature learning model with more robust feature learning abilities and better interpretability [46, 49]. For DIFL-FR, the parameters of the TSK-FS are optimized by a specific objective function, which enables the interpretation of the feature extraction model using the rules. Third, it is a deep learning system where a layer-wise strategy is used to learn the features from the data automatically and efficiently.

The main contributions of the paper are summarized as follows:

1) Different from the classic TSK-FS that is usually used for classification and regression, it is developed as a feature extraction model for image feature learning in this work.

2) By using the stacked structure and sliding window strategy of deep learning, a robust deep TSK-FS image feature learning method against noise and occlusion in images is realized.

3) Extensive experiments are conducted on image datasets of different scales. The experimental results show clearly the effectiveness and superiority of the proposed method.

The remainder of this paper is organized as follows. The related work is given in Section II. The details of the proposed method are discussed in Section III. In Section IV, the experiments are presented and the results are analyzed. Finally, the conclusions are given in Section V.

## II. Related work

In our method, sliding window strategy used in CNN, stacked structure and TSK-FS are integrated to implement unsupervised fuzzy-rule-based deep feature learning. The relevant background knowledge of CNN for image feature learning is first reviewed. Then, classical unsupervised deep learning methods based on Restricted Boltzmann Machine (RBM) [43] and autoencoder [44] are introduced, followed by a brief description of the fundamentals of TSK-FS.

### *A. Convolution Neural Network*

As one of the classical deep learning models, CNNs [17, 25–27] have been the most widely used structures in the field of image processing. CNNs are mainly composed of convolutional layers, pooling layers and fully connected layers. Convolutional layer is the core of the CNNs, which imitates the mechanism of the local receptive field in human vision. Different convolutional kernels can extract different features, such as edges, textures or corners. In general, the outputs of the convolutional layer are activated by nonlinear functions, and the feature maps are then formed by the activated results. The commonly used activation functions include the sigmoid function and the ReLU function. The pooling layer, also called the subsampling layer, conducts partial downsampling on the feature maps from the previous layer. The commonly used methods include maximum pooling and average pooling. The complexity of the model can be reduced by the pooling operations, which also results in insensitivity to translation and rotation of the images. The fully connected layer is equivalent to the hidden layer in traditional feedforward neural network. It is usually built at the end of the CNN. The purpose of the fully connected layer is to map the features learned by the network into the label space of the samples. In some CNNs, the function of the fully connected layer can be partially replaced by global average pooling [28].

### *B. Classical Unsupervised Deep Feature Learning Models*

Restricted Boltzmann Machine (RBM) [43] is a classical unsupervised feature learning method. It is a kind of stochastic neural network that can learn the probability distribution from the input data. Deep Boltzmann Machine (DBM) [40] is a deep learning model based on RBM, which is composed of multiple layers of RBM. Deep belief network (DBN) [17] is also a neural network based on RBM, which is composed of several layers of neurons, with RBM as the components. DBN can be used for both unsupervised learning and supervised learning. Regardless of the learning paradigms, the essence of DBN is the process of feature learning, that is, how to obtain a better feature representation.

Autoencoder is another representative neural network for unsupervised feature learning. Its structure is divided into two parts: encoder and decoder. The hidden layer feature outputs produced by the encoder, that is, the "encoded features" can be regarded as a representation of the input data. In the early stage of the research, autoencoder was used to solve the "encoder problem" in representation learning. Deep autoencoder [47] was also proposed for dimensionality reduction, which has shown better performance than many traditional methods, such as PCA [13].

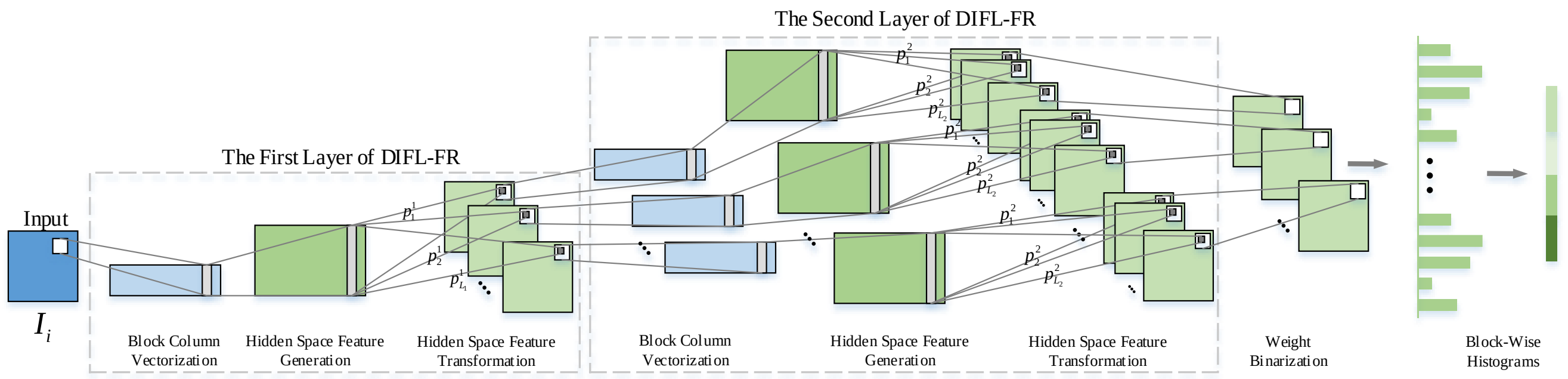

Fig. 1. The architecture of the proposed DIFL-FR.

## C. TSK Fuzzy System

Fuzzy system [29] is a model based on fuzzy rules and fuzzy logic. The fuzzy rules are generally obtained in two ways, designed based on expert experience, or derived from model training on the data. By using fuzzy sets [30], fuzzy system can directly transform natural semantics of human into machine languages recognizable by computers. Due to powerful learning ability and good interpretability, fuzzy systems are increasingly being applied to various fields, such as pattern recognition, intelligent control, data mining, and image processing. TSK-FS [29, 31] is one of the most popular fuzzy system models. In this paper, we explore the image feature learning ability of TKS-FS.

TSK-FS contains a fuzzy rule base. In general, the $k$-th fuzzy rule of the rule base can be formulated as follows:

$$\begin{aligned} &\mathrm{R}^k\text{: IF } x_1 \text{ is } A_1^k \wedge x_2 \text{ is } A_2^k \wedge \ldots \wedge x_d \text{ is } A_d^k, \\ &\quad \text{THEN } f^k(\mathbf{x}) = p_0^k + p_1^k x_1 + \ldots + p_d^k x_d, \\ &\quad\quad k = 1, 2, \ldots, K \end{aligned} \tag{2.1}$$

where $\mathbf{x} = [x_1, x_2, \ldots, x_d]^{\mathrm{T}} \in R^{d\times 1}$ is an input vector, $A_i^k$ is the fuzzy set [32] for the $i$-th feature in the $k$-th rule, $\mathbf{p}^k = \left[p_0^k, p_1^k, \ldots, p_d^k\right]^{\mathrm{T}} \in R^{(d+1)\times 1}$ is the consequent parameter vector of the $k$-th rule, and $\wedge$ denotes the fuzzy conjunction operator. When specific fuzzy operations are adopted for TSK-FS, the final output of the model can be expressed as:

$$\begin{aligned} y &= \sum_{k=1}^{K} \frac{\mu^k(\mathbf{x})}{\sum_{k'}^{K} \mu^{k'}(\mathbf{x})} f^k(\mathbf{x}) = \sum_{k=1}^{K} \tilde{\mu}^k(\mathbf{x}) f^k(\mathbf{x}) \\ &= \sum_{k=1}^{K} \tilde{\mu}^k(\mathbf{x}) \left(p_0^k + p_1^k x_1 + \ldots + p_d^k x_d\right) \end{aligned} \tag{2.2a}$$

with

$$\mu^k(\mathbf{x}) = \prod_{i=1}^{d} \mu_{A_i^k}(x_i) \tag{2.2b}$$

$$\tilde{\mu}^k(\mathbf{x}) = \mu^k(\mathbf{x}) \Big/ \sum_{k'=1}^{K} \mu^{k'}(\mathbf{x}) \tag{2.2c}$$

where $\mu_{A_i^k}(x_i)$ is the membership of $x_i$ to fuzzy set $A_i^k$. If multiplication is used as the conjunction operator, the firing level of the $k$-th rule of each sample can be formulated as (2.2b), and its normalized form is expressed in (2.2c). Specific membership functions are defined for each fuzzy set in the fuzzy rules. In the paper, the Gaussian membership function is used:

$$\mu_{A_i^k}(x_i) = \exp\left(\frac{-\left(x_i - c_i^k\right)^2}{2\delta_i^k}\right) \tag{2.2d}$$

where $c_i^k$ and $\delta_i^k$ in (2.2d) are the antecedent parameters that can be estimated using different approaches, e.g., the fuzzy c-means clustering [33] or deterministic clustering [34] algorithms.

Once the antecedent parameters are obtained, TSK-FS can be represented as a linear model in a new feature space. The details are explained as follows:

$$\mathbf{x}_e = \left(1, \mathbf{x}^{\mathrm{T}}\right)^{\mathrm{T}} \in R^{(d+1)\times 1} \tag{2.3a}$$

$$\tilde{\mathbf{x}}^k = \tilde{\mu}^k(\mathbf{x})\mathbf{x}_e \in R^{(d+1)\times 1} \tag{2.3b}$$

$$\mathbf{x}_g = \left[\left(\tilde{\mathbf{x}}^1\right)^{\mathrm{T}}, \left(\tilde{\mathbf{x}}^2\right)^{\mathrm{T}}, \ldots, \left(\tilde{\mathbf{x}}^K\right)^{\mathrm{T}}\right]^{\mathrm{T}} \in R^{K(d+1)\times 1} \tag{2.3c}$$

$$\mathbf{p}^k = \left(p_0^k, p_1^k, \ldots, p_d^k\right)^{\mathrm{T}} \in R^{(d+1)\times 1} \tag{2.3d}$$

$$\mathbf{p}_g = \left[\left(\mathbf{p}^1\right)^{\mathrm{T}}, \left(\mathbf{p}^2\right)^{\mathrm{T}}, \ldots, \left(\mathbf{p}^K\right)^{\mathrm{T}}\right]^{\mathrm{T}} \in R^{K(d+1)\times 1} \tag{2.3e}$$

The output of the TSK-FS in (2.2a) can be expressed as:

$$y = \mathbf{p}_g^{\mathrm{T}} \mathbf{x}_g \tag{2.3f}$$

where $\mathbf{x}_g \in R^{K(d+1)\times 1}$ represents the feature vector in the new feature space transformed from the original input vector $\mathbf{x} \in R^{d\times 1}$, and $\mathbf{p}_g$ is the combination vector of the consequent parameters of all the fuzzy rules.

For TSK-FS with multiple outputs, the outputs can be expressed:

$$\mathbf{y} = \mathbf{P}_g^{\mathrm{T}} \mathbf{x}_g \tag{2.3g}$$

where $\mathbf{y} \in R^{L\times 1}$ is the output vector, $\mathbf{P}_g = [\mathbf{p}_g^1, \cdots, \mathbf{p}_g^L] \in R^{K(d+1)\times L}$ is the consequent parameter matrix, and $\mathbf{p}_g^i$ is the combination vector of the consequent parameters of all the fuzzy rules for the $l$ -th output. Details of a multi-output TSK-FS can be found in Part C of the *Supplementary Materials* section.

TABLE I
KEY NOTATIONS

| Notation | Description |
|---|---|
| $K$ | The number of fuzzy rules. |
| $\mathbf{x}=[x_1,x_2,...,x_d]^{\mathrm{T}}$ | The input vector of TSK fuzzy system, where $d$ is the length of $\mathbf{x}$ . |
| $A_i^k$ | The fuzzy set in the $k$-th rule corresponding to the $i$-th input $x_i$ in $\mathbf{x}$ |
| $\mu_{A_i^k}(x_i)$ | The membership of $x_i$ to fuzzy set $A_i^k$ . |
| $\mathbf{p}^k=\left[p_0^k,p_1^k,...,p_d^k\right]^{\mathrm{T}}$ | The consequent parameter vector of the $k$-th rule. |
| $\mathbf{p}_g=\left[(\mathbf{p}^1)^{\mathrm{T}},(\mathbf{p}^2)^{\mathrm{T}},...,(\mathbf{p}^K)^{\mathrm{T}}\right]^{\mathrm{T}}$ | The combination vector of the consequent parameters of all the fuzzy rules. |
| $\mathbf{P}_g=[\mathbf{p}_g^1,\cdots,\mathbf{p}_g^L]$ | The consequent parameter matrix corresponding to $L$ outputs. |
| $\mathbf{x}_g$ | the feature vector in the new feature space transformed by fuzzy rule antecedents from $\mathbf{x}$ . |
| $\mathbf{y}=\mathbf{P}_g^{\mathrm{T}}\mathbf{x}_g$ | The output vector of TSK fuzzy system |
| $\mathbf{I}=[\mathbf{I}_1,\mathbf{I}_2,\cdots,\mathbf{I}_N]$ | The training image dataset, where the input image $\mathbf{I}_i$ is of size $m\times n$ |
| $\mathbf{X}_i=[\mathbf{x}_{i,1},\mathbf{x}_{i,2},\cdots,\mathbf{x}_{i,mn}]$ | The set of vectorized patches in image $\mathbf{I}_i$ . |
| $\mathbf{g}_{i,j}=\mathbf{x}_{gi,j}$ | The feature vector in the new feature space transformed by fuzzy rule antecedents from the vectorized patch $\mathbf{x}_{i,j}$ in image $\mathbf{I}_i$ . |
| $\mathbf{G}_i=\left[\mathbf{g}_{i,1},\mathbf{g}_{i,2},\cdots,\mathbf{g}_{i,mn}\right]$ | The matrix formed by concatenating $\mathbf{g}_{i,j}$ in image $\mathbf{I}_i$ . |
| $\overline{\mathbf{G}}_i=[\overline{\mathbf{g}}_{i,1},\overline{\mathbf{g}}_{i,2},\cdots,\overline{\mathbf{g}}_{i,mn}]$ | The matrix obtained by centralizing the $\mathbf{G}_i$ . |
| $\overline{\mathbf{G}}^s=[\overline{\mathbf{G}}_1^s,\overline{\mathbf{G}}_2^s,\cdots,\overline{\mathbf{G}}_N^s]$ | The centralization matrix in the $s$-th layer of the proposed DIFL-FR. |
| $\mathbf{P}^s=[\mathbf{p}_g^{s,1},\mathbf{p}_g^{s,2},\cdots,\ \mathbf{p}_g^{s,L_s}]$ | The consequent parameters in the $s$-th layer of the proposed DIFL-FR. |
| $\mathbf{Z}^s=(\mathbf{P}^s)^{\mathrm{T}}\overline{\mathbf{G}}^s$ | The output vectors of the $s$-th layer of the proposed DIFL-FR. |
| $z_{i,l_s,\ell_{s-1},j}^s$ | The output of the $j$-th block in the $\ell_{s-1}$-th feature image of image $\mathbf{I}_i$ . |
| $\mathbf{I}_{i,l_s,\ell_{s-1}}^s$ | The feature image combined by $mn$ feature values $z_{i,l_s,\ell_{s-1},j}^s$ for image $\mathbf{I}_i$ . |
| $\mathbf{P}_{i,l_s,\ell_{s-1}}=H\left(\mathbf{I}_{i,l_s,\ell_{s-1}}^s\right)$ | the feature image outputted by the binarization function. |
| $\mathbf{T}_{i,\ell_{s-1}}=\sum_{l_s=1}^{L_s}2^{l_s-1}\mathbf{P}_{i,l_s,\ell_{s-1}}$ | The integer-valued feature image converted by $\mathbf{P}_{i,l_s,\ell_{s-1}}$ . |
| $\mathbf{h}_{i,\ell_{s-1}}=Hist\left(\mathbf{T}_{i,\ell_{s-1}}\right)$ | The vector concatenated by the block histogram statistics. |
| $\mathbf{f}_i=\left[\mathbf{h}_{i,1};\mathbf{h}_{i,2};\cdots;\mathbf{h}_{i,\Gamma_{s-1}}\right]$ | The final output feature vector of the original input image $\mathbf{I}_i$ . |

**Remark**: TSK-FS has been widely used for classification and regression. For example, a rule-based interpretable and discriminative model can be obtained for classification by using labeled datasets and supervised learning methods to train the TSK-FS system. Instead, this paper investigates the atypical use of TSK-FS on image feature extraction.

## III. DEEP IMAGE FEATURE LEARNING WITH FUZZY RULES

The proposed DIFL-FR model is a cascaded structure consisting of three components: multilayer TSK-FS image feature learning, weight binarization, and blockwise histograms. It realizes nonlinear transformation by using the antecedent part of the multi-output TSK-FS to generate hidden features. The transformation has good interpretability and a nonlinear feature learning ability similar to that of the activation functions in classic CNNs. It also. DIFL-FR generates different new features with multi-group consequent parameters that are similar to the convolution kernels in CNNs. The multilayer TSK-FS image feature learning can extract deeper image features in a progressive way.

The proposed DIFL-FR is discussed in five parts. The overall structure of DIFL-FR is presented in Section III-A. The first and the subsequent layers of the TSK-FS image feature learning are discussed in Sections III-B and III-C, respectively. The details of the output layer are given in Section III-D. The key notations are listed in Table I.

### *A. The Architecture of DIFL-FR*

The proposed DIFL-FR is an end-to-end learning method which learns features automatically from raw data. It introduces fuzzy set to represent the information and reduces the sensitivity to noise. Meanwhile, multilayer TSK-FS with stacked structures are integrated to implement feature learning, where fuzzy rules and fuzzy inference are used to model human decision process and extract highly discriminative features. The architecture of DIFL-FR contains two layers of TSK-FSs, as illustrated in Fig. 1.

### *B. The First Layer of DIFL-FR*

The TSK-FS image feature learning process has three steps: block column vectorization with patch sliding, feature generation in the hidden space, and feature transformation in the hidden space. Fig. S1 in Part A of the *Supplementary Materials* section shows the process. First, TSK-FS image feature learning scans the original input image and vectorizes it by patch sliding and block column vectorization. Second, nonlinear transformation is implemented by the antecedent part of the multi-output TSK-FS to generate a hidden feature space. Third, feature dimensional reduction is achieved by linear transformation of the hidden space through the consequent part of the multi-output TSK-FS. The details are given as follows.

For model training with $N$ images and each with a size of $m \times n$, the training image dataset can be represented as $\{\mathbf{I}_i\}_{i=1}^N$. The $N$ images are concatenated as a matrix $\mathbf{I} = [\mathbf{I}_1, \mathbf{I}_2, \cdots, \mathbf{I}_N]$.

#### *1) Block Column Vectorization with Patch Sliding*

For the input image $\mathbf{I}_i$ of size $m \times n$, we use a patch of size $h_1 \times h_2$ to scan each pixel (the edges of the image are filled with 0) and then reshape each $h_1 \times h_2$ matrix into a column vector (as shown in Fig. S2 in Part A of the *Supplementary Materials* section). Then, all the vectors corresponding to image $\mathbf{I}_i$ can be represented as a matrix:

$$\mathbf{X}_i = [\mathbf{x}_{i,1}, \mathbf{x}_{i,2}, \cdots, \mathbf{x}_{i,mn}] \in R^{h_1 h_2 \times mn}, i = 1, 2, \cdots, N \tag{3.1}$$

where $\mathbf{x}_{i,j} \in R^{h_1 h_2 \times 1}$ denotes the $j$-th vectorized patch in image $\mathbf{I}_i$. Therefore, for all the training images $\{\mathbf{I}_i\}_{i=1}^N$, the vector set can be represented by the following matrix:

$$\mathbf{X} = [\mathbf{X}_1, \mathbf{X}_2, \cdots, \mathbf{X}_N] \in R^{h_1 h_2 \times mnN} \tag{3.2}$$

#### *2) Hidden Space Feature Generation*

Based on the principle of TSK-FS, the fuzzy membership in the antecedent part of the fuzzy rules can be generated according to (2.2b)-(2.2d). Clustering is a commonly used technique to estimate the antecedent parameters of the TSK-FS. Although many models such as K-means and Gaussian mixture model (GMM) are proposed to solve clustering problems, they depend on the initial guess of the partitions. Var-Part [42] is a deterministic clustering algorithm which can avoid randomized initializations and is computationally efficient. It has been proved that Var-Part can yield sum squared-error values close to the optimum values obtained by several random-start with K-means [42]. This improves the quality of the antecedent parameters of the TSK-FS. Therefore, we adopt Var-Part in the proposed method DIFL-FR. The details of Var-Part are given in Part B of the *Supplementary Materials* section.

For the original image $\mathbf{I}_i$, the original dataset $\mathbf{X}_i = [\mathbf{x}_{i,1}, \mathbf{x}_{i,2}, \cdots, \mathbf{x}_{i,mn}] \in R^{h_1 h_2 \times mn}$ can be constructed through block column vectorization, where $\mathbf{x}_{i,j} \in R^{h_1 h_2 \times 1}$ denotes the $j$-th vectorized patch in image $\mathbf{I}_i$. Once the fuzzy membership is determined, according to (2.3a)-(2.3c), the original dataset generated from the $N$ images can be expressed as $\mathbf{X} = [\mathbf{X}_1, \mathbf{X}_2, \cdots, \mathbf{X}_N] \in R^{h_1 h_2 \times mnN}$. Then, $\mathbf{X}$ can be mapped from the original feature space to the hidden feature space, and the dataset $\mathbf{G}$ in the new hidden feature space can be obtained as follows based on the mechanism in Eqs. (2.3a)-(2.3c).

$$\mathbf{G} = [\mathbf{G}_1, \mathbf{G}_2, \cdots, \mathbf{G}_N] \in R^{K(h_1 h_2 + 1) \times mnN} \tag{3.3a}$$

$$\begin{aligned} \mathbf{G}_i &= \left[\mathbf{g}_{i,1}, \mathbf{g}_{i,2}, \cdots, \mathbf{g}_{i,mn}\right] \\ &= \left[\mathbf{x}_{gi,1}, \mathbf{x}_{gi,2}, \cdots, \mathbf{x}_{gi,mn}\right] \in R^{K(h_1 h_2 + 1) \times mn} \end{aligned} \tag{3.3b}$$

$$\mathbf{g}_{i,j} = \mathbf{x}_{gi,j} \in R^{K(h_1 h_2 + 1) \times 1} \tag{3.3c}$$

where $\mathbf{G}$ is the concatenated dataset of $\mathbf{G}_i$ for all the images after the conversion of dataset $\mathbf{X}_i$ in original feature space into the dataset $\mathbf{G}_i$ in the new hidden feature space. $\mathbf{G}_i$ is a matrix formed by concatenating $\mathbf{g}_{i,j}$, where $\mathbf{g}_{i,j} = \mathbf{x}_{gi,j}$ as defined by (2.3a)-(2.3c). An illustration about the process is presented in Fig. S3 in Part A of the *Supplementary Materials* section.

#### *3) Hidden Space Feature Transformation*

If the data transformed by the antecedent part of the multi-output TSK-FS is viewed as a feature representation in the new hidden space, the consequent part of the multi-output TSK-FS can be viewed as a linear dimensional reduction in this new space as shown in (2.3g). To preserve the geometric properties of the data during dimensional reduction, the criterion of PCA is applied by maximizing the variance of the data in the hidden feature space to optimize the consequent parameters $\mathbf{P}$ of the TSK-FS.

When the criterion of PCA is adopted to train the consequent parameters of the TSK-FS for feature learning, it is essential that we implement PCA on the dataset $\mathbf{G}$ in the hidden feature space which is transformed from the dataset $\mathbf{X}$ in the original feature space. Since the outputs of the trained TSK-FS can be expressed as $\mathbf{P}^{\mathrm{T}}\mathbf{G}$, each column vector in the consequent parameter matrix $\mathbf{P}$ just corresponds to an eigenvector of PCA. According to the principle of PCA, the optimization objective of the TSK-FS for feature learning can then be formulated in the same form of PCA, i.e.,

$$\max_{\mathbf{P}} Tr(\mathbf{P}^{\mathrm{T}}\overline{\mathbf{G}}\,\overline{\mathbf{G}}^{\mathrm{T}}\mathbf{P}), s.t. \mathbf{P}^{\mathrm{T}}\mathbf{P} = \mathbf{I}_{L_1} \quad (3.4a)$$

where $\overline{\mathbf{G}}$ is the matrix obtained by centralizing the hidden feature space data matrix $\mathbf{G}$, as in the classic PCA, and $\mathbf{I}_{L_1}$ is an identity matrix of size $L_1 \times L_1$ with $L_1$ being the number of outputs of the TSK-FS, i.e., the dimension of the output features. Specifically, for each training image $\mathbf{I}_i$, the following centralization matrix can be obtained:

$$\overline{\mathbf{G}}_i = [\overline{\mathbf{g}}_{i,1}, \overline{\mathbf{g}}_{i,2}, \cdots, \overline{\mathbf{g}}_{i,mn}] \in R^{K(h_1 h_2+1)\times mn} \quad (3.4b)$$

where $\overline{\mathbf{g}}_{i,j} = \mathbf{g}_{i,j} - \frac{1}{N}\sum_{i'=1}^{N}\sum_{j'=1}^{mn}\mathbf{g}_{i',j'}$ .Thus, for all the training images $\{\mathbf{I}_i\}_{i=1}^{N}$, the centralization matrix of the dataset can be obtained by $\overline{\mathbf{G}} = [\overline{\mathbf{G}}_1, \overline{\mathbf{G}}_2, \cdots, \overline{\mathbf{G}}_N] \in R^{K(h_1 h_2+1)\times mnN}$ .

The Lagrange multiplier method can be used to transform (3.4a) to the following equivalent problem.

$$\overline{\mathbf{G}}\,\overline{\mathbf{G}}^{\mathrm{T}}\mathbf{p}_g^i = \lambda_i^1 \mathbf{p}_g^i \quad (3.5a)$$

Therefore, the optimization problem in (3.4a) can be further transformed into the eigenvalue decomposition problem below:

$$\mathbf{C}_1 = \mathbf{P}\mathbf{\Lambda}_1\mathbf{P}^{\mathrm{T}} \quad (3.5b)$$

with

$$\mathbf{C}_1 = \frac{1}{mnN}\overline{\mathbf{G}}\,\overline{\mathbf{G}}^{\mathrm{T}} \in R^{K(h_1 h_2+1)\times K(h_1 h_2+1)} \quad (3.5c)$$

$\mathbf{\Lambda}_1$ is a diagonal matrix composed of the first $L_1$ largest eigenvalues of $\mathbf{C}_1$, i.e.,

$$\mathbf{\Lambda}_1 = \mathrm{diag}(\lambda_{l_1}^{1}),\ l_1 = 1, 2, \cdots, L_1 \quad (3.5d)$$

where $\lambda_1^1 \ge \lambda_2^1 \ge \cdots \ge \lambda_{L_1}^1$ . $\mathbf{P}$ is then given by the matrix composed of the corresponding eigenvectors as follows:

$$\mathbf{P} = [\mathbf{p}_g^1, \mathbf{p}_g^2, \cdots,\ \mathbf{p}_g^{L_1}] \in R^{K(h_1 h_2+1)\times L_1} \quad (3.5e)$$

where $\mathbf{p}_g^{l_1}$ represents the $l_1$-th eigenvector, which is also the consequent parameter vector corresponding to the $l_1$-th output of the multi-output TSK-FS.

According the principle of PCA, the outputs of the TSK-FS, i.e., $\mathbf{P}^{\mathrm{T}}\mathbf{G}$, is modified as $\mathbf{P}^{\mathrm{T}}\overline{\mathbf{G}}$ to get the extracted new features. Once the consequent parameters $\mathbf{P}$ are determined, the new feature data learned by TSK-FS can be obtained as follows:

$$\mathbf{Z}^1 = \mathbf{P}^{\mathrm{T}}\overline{\mathbf{G}} \quad (3.6a)$$

$$\mathbf{Z}^1 = [\mathbf{Z}_1^1; \mathbf{Z}_2^1; \cdots; \mathbf{Z}_{L_1}^1] \quad (3.6b)$$

$$\mathbf{Z}_{l_1}^1 = [\mathbf{z}_{1,l_1}^1, \mathbf{z}_{2,l_1}^1, \cdots, \mathbf{z}_{N,l_1}^1],\ l_1 = 1, 2, \cdots, L_1 \quad (3.6c)$$

$$\mathbf{z}_{i,l_1}^1 = [z_{i,l_1,1}^1, z_{i,l_1,2}^1, \cdots, z_{i,l_1,mn}^1],\ i = 1, 2, \cdots, N \quad (3.6d)$$

$$z_{i,l_1,j}^1 = \left(\mathbf{p}_g^{l_1}\right)^{\mathrm{T}}\overline{\mathbf{g}}_{i,j}, \quad i = 1, 2, \cdots, N;\ l_1 = 1, 2, \cdots, L_1;\ j = 1, 2, \cdots, mn \quad (3.6e)$$

Here, $z_{i,l_1,j}^1$ is the feature of the $j$-th block in the $i$th image, which is obtained by the $l_1$-th group of the consequent parameters for the $l_1$-th output. For each training image $\mathbf{I}_i$, the number of features obtained by the $l_1$-th group of the consequent parameters is $mn$. These results can be reconstructed as an image of the same size as the original training image. Since there are $L_1$ groups of consequent parameters in the first layer, for each image we have $L_1$ new feature images. Thus, the feature image of the first layer is formed as (3.7).

$$\mathbf{I}_{i,l_1}^1 \in R^{m\times n}, l_1 = 1, 2, \cdots, L_1 \quad (3.7)$$

For all the input training images $\{\mathbf{I}_i\}_{i=1}^{N}$, the matrix corresponding to the set of feature images can be expressed as:

$$\mathbf{I}^1 = [\mathbf{I}_{1,1}^1, \mathbf{I}_{1,2}^1, \cdots, \mathbf{I}_{1,L_1}^1, \cdots, \mathbf{I}_{N,1}^1, \mathbf{I}_{N,2}^1, \cdots, \mathbf{I}_{N,L_1}^1] \quad (3.8)$$

The PCA optimization criterion adopted in this paper is just a viable option. Other more sophisticated criteria can also be used to optimize the consequent parameters of the TSK-FS for feature learning, which will be investigated in our future work.

## *C. The Subsequent Layers of DIFL-FR*

The process of constructing the subsequent $s$-th layer $(s \ge 2)$ of DIFL-FR is basically the same as that of the first layer of DIFL-FR as presented above in Subsection III-B. The steps are summarized as follows.

*1) Block column vectorization*

Refer to the steps of block column vectorization in the first layer and use the output of the ($s$-1)-th layer as the input to the $s$-th layer, the feature image is given by $\mathbf{I}_{i,\ell_{s-1}}^{s-1},\ i = 1, 2, \cdots, N,\ \ell_{s-1} = 1, 2, \cdots, \Gamma_{s-1},\ \Gamma_{s-1} = \prod_{n_s=1}^{s-1} L_{n_s}$ , where each pixel point has a patch of size $h_1 \times h_2$, and the information of all the points in the patch is concatenated through block vectorization. Finally, the following data matrix can be obtained based on the feature image $\mathbf{I}_{i,\ell_{s-1}}^{s-1}$ :

$$\mathbf{X}_{i,\ell_{s-1}}^{s} = [\mathbf{x}_{i,\ell_{s-1},1}^{s}, \mathbf{x}_{i,\ell_{s-1},2}^{s}, \cdots, \mathbf{x}_{i,\ell_{s-1},mn}^{s}] \in R^{h_1 h_2\times mn} \quad (3.9)$$

where $\mathbf{I}_{i,\ell_{s-1}}^{s-1}$ represents the $\ell_{s-1}$-th feature image of the $i$-th original image obtained through the first $s$-1 layer TSK-FS image feature learning, and $\mathbf{x}_{i,\ell_{s-1},j}^{s} \in R^{h_1 h_2\times 1}$ represents the column vector of the $j$-th block in $\mathbf{I}_{i,\ell_{s-1}}^{s-1}$ . Therefore, for each original image $\mathbf{I}_i$, the obtained total column vectors in the $s$-th $(s \ge 2)$ layer can be represented as follows:

$$\mathbf{X}_i^s = [\mathbf{X}_{i,1}^s, \mathbf{X}_{i,2}^s, \cdots, \mathbf{X}_{1,\Gamma_{s-1}}^s] \in R^{h_1 h_2\times \Gamma_{s-1} mn} \quad (3.10)$$

For all the original training image set $\{\mathbf{I}_i\}_{i=1}^{N}$, the matrices $\mathbf{X}_i^s,\ i = 1, 2, \cdots, N,$ can be concatenated to yield the total column vectors in (3.11) that are used to train the consequent parameters of the TSK-FS in the $s$-th layer of DIFL-FR.

$$\mathbf{X}^s = [\mathbf{X}_1^s, \mathbf{X}_2^s, \cdots, \mathbf{X}_N^s] \in R^{h_1 h_2 \times \Gamma_{s-1} mnN} \quad (3.11)$$

*2) Hidden Space Feature Generation*

Similar to the hidden space feature generation in the first layer of DIFL-FR, the input data in the *s*-th layer $\mathbf{X}^s$ can be mapped to the new hidden feature space, and the corresponding data in the new feature space can be expressed as:

$$\mathbf{G}^s = [\mathbf{G}_1^s, \mathbf{G}_2^s, \cdots, \mathbf{G}_N^s] \in R^{K_s(h_1 h_2+1) \times \Gamma_{s-1} mnN}$$
$$\Gamma_{s-1} = \prod_{n_s=1}^{s-1} L_{n_s} \quad (3.12a)$$

$$\mathbf{G}_i^s = [\mathbf{G}_{i,1}^s, \mathbf{G}_{i,2}^s, \cdots, \mathbf{G}_{i,\Gamma_{s-1}}^s] \in R^{K_s(h_1 h_2+1) \times \Gamma_{s-1} mn}, \quad i = 1, 2, \cdots, N \quad (3.13b)$$

$$\mathbf{G}_{i,\ell_{s-1}}^s = [\mathbf{g}_{i,\ell_{s-1},1}^s, \mathbf{g}_{i,\ell_{s-1},2}^s, \cdots, \mathbf{g}_{i,\ell_{s-1},mn}^s] \in R^{K_s(h_1 h_2+1) \times mn}, \quad \ell_{s-1} = 1, 2, \cdots, \Gamma_{s-1} \quad (3.13c)$$

where $\mathbf{g}_{i,\ell_{s-1},j}^s$ is defined in (3.3c), and $K_s$ is the number of fuzzy rules in the *s*-th layer of the TSK-FS in the DIFL-FR.

*3) Hidden Space Feature Transformation*

By centralizing the hidden space data $\mathbf{G}^s$, we get

$$\overline{\mathbf{G}}^s = [\overline{\mathbf{G}}_1^s, \overline{\mathbf{G}}_2^s, \cdots, \overline{\mathbf{G}}_N^s] \in R^{K_s(h_1 h_2+1) \times \Gamma_{s-1} mnN}, \quad \Gamma_{s-1} = \prod_{n_s=1}^{s-1} L_{n_s} \quad (3.13d)$$

$$\overline{\mathbf{G}}_i^s = [\overline{\mathbf{G}}_{i,1}^s, \overline{\mathbf{G}}_{i,2}^s, \cdots, \overline{\mathbf{G}}_{i,\Gamma_{s-1}}^s] \in R^{K_s(h_1 h_2+1) \times \Gamma_{s-1} mn}, \quad i = 1, 2, \cdots, N \quad (3.13e)$$

$$\overline{\mathbf{G}}_{i,\ell_{s-1}}^s = [\overline{\mathbf{g}}_{i,\ell_{s-1},1}^s, \overline{\mathbf{g}}_{i,\ell_{s-1},2}^s, \cdots, \overline{\mathbf{g}}_{i,\ell_{s-1},mn}^s] \in R^{K_s(h_1 h_2+1) \times mn}, \quad \ell_{s-1} = 1, 2, \cdots, \Gamma_{s-1} \quad (3.13f)$$

where $\overline{\mathbf{g}}_{i,\ell_{s-1},j}^s$ is a mean-removed vector as defined in (3.4b).

Similar to the first layer of DIFL-FR, the PCA optimization criterion is used to solve for the consequent parameters $\mathbf{P}^s = [\mathbf{p}_g^{s,1}, \mathbf{p}_g^{s,2}, \cdots, \mathbf{p}_g^{s,L_s}] \in R^{K(h_1 h_2+1) \times L_s}$ of the TSK-FS in the *s*-th layer of the DIFL-FR. Here, $\mathbf{p}_g^{s,l_s}$ represents the consequent parameter corresponding to the $l_s$-th output of the multi-output TSK-FS.

Once the consequent parameters $\mathbf{P}^s$ are determined, the new feature learned by the *s*-th layer of DIFL-FR can be obtained:

$$\mathbf{Z}^s = \left(\mathbf{P}^s\right)^{\mathrm{T}} \overline{\mathbf{G}}^s \quad (3.14a)$$

$$\mathbf{Z}^s = [\mathbf{Z}_1^s; \mathbf{Z}_2^s; \cdots; \mathbf{Z}_{L_s}^s] \quad (3.14b)$$

$$\mathbf{Z}_{l_s}^s = [\mathbf{Z}_{1,l_s}^s, \mathbf{Z}_{2,l_s}^s, \cdots, \mathbf{Z}_{N,l_s}^s], \quad l_s = 1, 2, \cdots, L_s \quad (3.14c)$$

$$\mathbf{Z}_{i,l_s}^s = [\mathbf{z}_{i,l_s,1}^s, \mathbf{z}_{i,l_s,2}^s, \cdots, \mathbf{z}_{i,l_s,\Gamma_{s-1}}^s], \quad i = 1, 2, \cdots, N, \quad \Gamma_{s-1} = \prod_{n_s=1}^{s-1} L_{n_s} \quad (3.14d)$$

$$\mathbf{z}_{i,l_s,\ell_{s-1}}^s = \left[z_{i,l_s,\ell_{s-1},1}^s, z_{i,l_s,\ell_{s-1},2}^s, \cdots, z_{i,l_s,\ell_{s-1},mn}^s\right], \quad \ell_{s-1} = 1, 2, \cdots, \Gamma_{s-1} \quad (3.14e)$$

$$z_{i,l_s,\ell_{s-1},j}^s = \left(\mathbf{p}_{\mathbf{g}}^{s,l_s}\right)^{\mathrm{T}} \overline{\mathbf{g}}_{i,\ell_{s-1},j}^s, \; j = 1, 2, \cdots, mn \quad (3.14f)$$

where $z_{i,l_s,\ell_{s-1},j}^s$ is the new feature of the *j*-th block in the $\ell_{s-1}$-th feature image $\mathbf{I}_{i,\ell_{s-1}}^{s-1}$, which is obtained by the $l_s$-th group of the consequent parameters of the TSK-FS in the *s*-th layer. $\mathbf{I}_{i,\ell_{s-1}}^{s-1}$ is the $\ell_{s-1}$-th feature image obtained after the *i*-th original image is learned by the TSK-FSs in the first *s-1* layers of the DIFL-FR. For each feature image $\mathbf{I}_{i,\ell_{s-1}}^{s-1}$, a total of *mn* new feature values $z_{i,l_s,\ell_{s-1},j}^s$ can be obtained. These results can be reconstructed into an image of the same size as the original training image. Thus, the feature image $\mathbf{I}_{i,l_s,\ell_{s-1}}^s$ is formed. There are $L_s$ groups of consequent parameters in the *s*-th layer, so that each feature image $\mathbf{I}_{i,\ell_{s-1}}^{s-1}$ can generate $L_s$ new feature images. For each original image, a total of $\Gamma_s$ feature images can be obtained with $\Gamma_s = \prod_{n_s=1}^{s} L_{n_s}$. Therefore, for all the training images, $\{\mathbf{I}_i\}_{i=1}^N$, after *s* layers of TSK-FS image feature learning, the feature image set can be represented as:

$$\left\{\mathbf{I}_{i,\ell_s}^s \in R^{m \times n}\right\}, \; i = 1, 2, \cdots, N; \ell_s = 1, 2, \cdots, \Gamma_s; \Gamma_s = \prod_{n_s=1}^{s} L_{n_s}$$

The feature image set $\mathbf{I}^s$ can be expressed in the matrix form as follows:

$$\mathbf{I}^s = [\mathbf{I}_{1,1}^s, \mathbf{I}_{1,2}^s, \cdots, \mathbf{I}_{1,\Gamma_S}^s, \cdots, \mathbf{I}_{N,1}^s, \mathbf{I}_{N,2}^s, \cdots, \mathbf{I}_{N,\Gamma_s}^s], \quad \Gamma_s = \prod_{n_s=1}^{s} L_{n_s} \quad (3.15)$$

where $n_s$ is the number of outputs in the $s$-th TSK-FS of DIFL-FR.

*D. The Output Layer*

The output layer of DIFL-FR first performs weight binarization on the feature image extracted by the previous cascaded TSK-FSs and then converts it into a block histogram statistical vector as the final feature extracted by the model.

*1) Weight Binarization*

To binarize the feature images obtained by the *s*-th layer, let $\mathbf{I}_{i,l_s,\ell_{s-1}}^s$ ($l_s = 1, 2, \cdots, L_s$) be the $l_s$-th feature image generated by the *s*-th layer TSK-FS based on the input feature image $\mathbf{I}_{i,l_{s-1}}^{s-1}$ ($\ell_{s-1} = 1, 2, \cdots, \Gamma_{s-1}$), the binarization function is defined as:

$$\mathbf{P}_{i,l_s,\ell_{s-1}} = H\left(\mathbf{I}_{i,l_s,\ell_{s-1}}^s\right), \; i = 1, 2, \cdots, N; \quad l_s = 1, 2, \cdots, L_s; \ell_{s-1} = 1, 2, \cdots, \Gamma_{s-1} \quad (3.16)$$

where $H(x) = \begin{cases} 1, & x \ge 0 \\ 0, & x < 0 \end{cases}$ is a Heaviside step function.

Thus, these $L_s$ binary images can be converted into an integer-valued feature image denoted as $\mathbf{T}_{i,\ell_{s-1}}$, which represents an integer-valued feature image generated by the $\ell_{s-1}$-th input feature image. It is obtained from the *i*-th original input image in the (*s-1*)-th layer TSK-FS. A pixel in $\mathbf{T}_{i,\ell_{s-1}}$ is an integer in the range $\left[0, 2^{L_s} - 1\right]$.

TABLE II
THE COMPUTATIONAL COMPLEXITY OF DIFFERENT ALGORITHMS

| Method | Computational Complexity |
|---|---|
| Raw | $O(1)$ |
| BlockHist | $O(m\times n)$ |
| LBP [2] | $O(8\times m\times n)$ |
| PCA [13] | $O((m\times n)^3)$ |
| CNN | $O(k\times(m\times n)\times d^2)$ |
| RBM [43] | $O(d\times(m\times n)^2)$ |
| Autoencoder [44] | $O(2\times d(m\times n))$ |
| DIFL-FR | $O(K\times(m\times n)^2+d(K\times m\times n))$ |

$$\mathbf{T}_{i,\ell_{s-1}}=\sum_{l_s=1}^{L_s}2^{l_s-1}\mathbf{P}_{i,l_s,\ell_{s-1}} \tag{3.17}$$

All the integer-valued feature images corresponding to all the original input images can be expressed by the following matrix: $\mathbf{T}=\left[\mathbf{T}_{1,1},\mathbf{T}_{1,2},\cdots,\mathbf{T}_{1,\Gamma_{S-1}},\cdots,\mathbf{T}_{N,1},\mathbf{T}_{N,2},\cdots,\mathbf{T}_{N,\Gamma_{S-1}}\right]$.

*2) Blockwise Histograms*

We use a block of size $h_1\times h_2$ to slide the integer-valued feature image $\mathbf{T}_{i,\ell_{s-1}}$ with an overlap ratio of $Cr$ (the default value is 0.5, meaning that the step sizes in two directions are $\lfloor h_1/2\rfloor$ and $\lfloor h_2/2\rfloor$, respectively), and partition $\mathbf{T}_{i,\ell_{s-1}}$ into $B$ blocks. For each block, the histogram (in the range $\left[0,2^{L_s}-1\right]$) of the decimal values can be computed, and the output of each block is a $2^{L_s}$ -dimensional vector. These $2^{L_s}$ dimensional vectors can be concatenated into a vector, expressed as $\mathbf{h}_{i,\ell_{S-1}}=Hist\left(\mathbf{T}_{i,\ell_{S-1}}\right)\in R^{2^{L_s}B\times 1}$, where $Hist(\ )$ is the operator that uses the block histogram statistics and expands the result as a vector.

After the *k*-layer image feature learning based on TSK-FS, weight binarization and blockwise histogram operations, the output feature vector $\mathbf{f}_i$ of the original input image $\mathbf{I}_i$ can be expressed as

$$\mathbf{f}_i=\left[\mathbf{h}_{i,1};\mathbf{h}_{i,2};\cdots;\mathbf{h}_{i,\Gamma_{s-1}}\right]\in R^{2^{L_s}B\Gamma_{s-1}\times 1},\quad \Gamma_{s-1}=\prod_{n_s=1}^{s-1}L_{n_s} \tag{3.18}$$

Finally, the extracted features can be put into a common classifier for learning. Table S1 in part D of the *Supplementary Materials* section provides a detailed algorithm description of the proposed DIFL-FR method.

### *E. Computational Complexity Analysis*

The computational complexity of the proposed DIFL-FR is analyzed and compared with some traditional methods. Table II compares the computational complexity of DIFL-FR with traditional methods. The handcraft-based methods are relatively simple and consume less time. RAW directly uses the original data as inputs and the complexity is $O(1)$. The complexity of BlockHist, related only to the input image size $m\times n$, is given by $O(m\times n)$. Based on BlockHist, LBP [2] compares the peripheral elements of the sliding window, and its complexity is $O(8\times m\times n)$. The computational complexity of PCA [13] is dependent on the complexity of covariance matrix computation and eigenvalue decomposition, i.e., $O((m\times n)^2)$ and $O((m\times n)^3)$ respectively, the overall complexity is $O((m\times n)^3)$. For learning-based methods, they usually have much higher computational complexity. The complexity of CNN is $O(k\times(m\times n)\times d^2)$, where $d$ is the length of output vectors, $k$ is the size of convolutional kernel. The complexity of the unsupervised learning methods RBM [43] and Autoencoder [44] is $O(d\times(m\times n)^2)$ and $O(2\times d(m\times n))$ respectively. For our method DIFL-FR, the complexity of using the clustering algorithm Var-Part to generate the antecedent part is $O(K\times(m\times n)^2)$. The complexity of solving the linear transformation in the consequent part is $O(d(K\times m\times n))$, $K$ is the number of fuzzy rules and also the number of clusters for clustering task. Hence, the overall complexity of DIFL-FR is $O(K\times(m\times n)^2+d(K\times m\times n))$, which is more efficient than the back-propagation learning-based methods.

## IV. EXPERIMENTS

Extensive experiments are conducted to evaluate the effectiveness of the proposed image extraction method DIFL-FR. The performance is evaluated with both small and large image datasets. In the experiments, unless otherwise specified, DIFL-FR with two layers is studied and multi-class linear SVM [35] is used as a classifier trained based on extracted image features. Classification accuracy is used as the indicator to evaluate the performance.

### *A. Small-Scale Datasets*

In this section, the proposed method DIFL-FR is evaluated on two small-scale face recognition datasets [36–38], i.e., the ORL and Extended Yale B datasets. Two handcraft-based methods, three matrix decomposition-based methods, and two neural networks-based methods are adopted as the comparative algorithms. The handcraft-based methods are the block histogram (BlockHist) and LBP [2] methods; the matrix decomposition-based methods are PCA [13], kernel principal component analysis (KPCA) and Fisherface [14] methods; and the neural networks-based methods are RBM [43] and autoencoder [44]. Deep learning methods are not considered since they usually perform poorly on small datasets. Meanwhile, the original images are treated as inputs of the benchmark algorithm "Raw", i.e., the original values of the pixels in the image are directly used by the classifier without feature extraction. Altogether, there are nine algorithms under comparison, including the proposed DILF-FR.

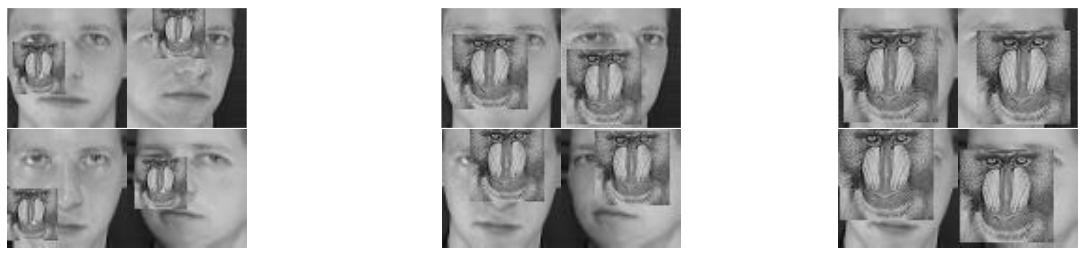

Fig. 3. Test face image with various levels of occlusion on the ORL dataset: (a) 20% occlusion, (b) 40% occlusion. (c) 60% occlusion.

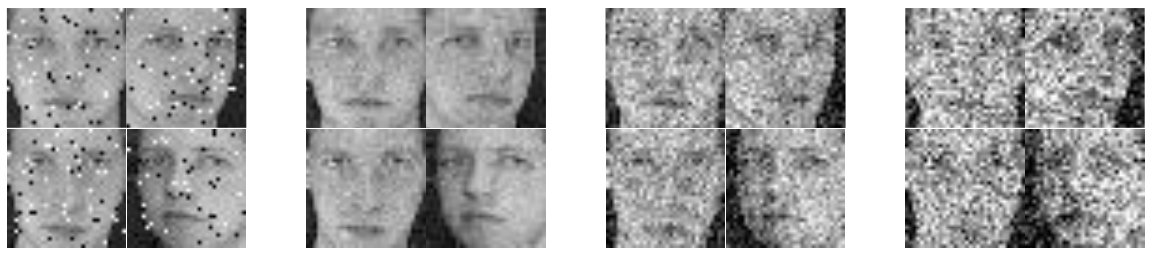

Fig. 2. Noise introduced to the ORL dataset (ORL_Tran_8): (a) salt & pepper noise, (b) (c) and (d) refer to the addition of Gaussian noise with 10, 30 and 50 standard deviations (std), respectively, to the training and test data of the ORL_Train_8 dataset.

In our experiments, all datasets are divided into training and test sets according to the proportions stated in Part E of the *Supplementary Materials* section. Each experiment is run for five times under the optimal parameter setting obtained from search grids of hyperparameters to get the mean accuracy. The details are given as follows. For the BlockHist method, the overlap ratio, $Cr$, is set as 0.5, and the block size is optimally set using the search grid $h \in \{3,5,7,9,11\}$. For LBP, the image is equally divided into $4\times4$ subareas. In each subarea, the histogram is counted using the uniform LBP, and then the obtained statistical histogram of each subarea is concatenated as a feature vector. For PCA and KPCA, the dimension of the subspace is optimally set using the search grid $m \in \{10,12,14,\cdots,400\}$. Gaussian kernel is used in KPCA, and the parameters of the Gaussian kernel are optimally set using the search grid $t \in \{2^{-4},2^{-3},\cdots,2^{4}\}$. For the Fisherface method, the dimension of the subspace is fixed to $C$-1 with $C$ being the number of classes. For RBM, the dimension of hidden units is optimally set using the search grid $d \in \{32,64,128,256,512\}$. The learning rate is set to 0.1, the number of training epochs is in the range $\{100,200,300,\cdots,1000\}$, and the training strategy of contrastive divergence is adopted to optimize model. For autoencoder, encoder and decoder with two layers are adopted, where the hidden layer structure is optimally set using the search grid $d \in \{32,64,128,256,512\}$. Meanwhile, the loss function of mean square error and the optimization algorithm of Adam are used for the autoencoder.

For the proposed DIFL-FR, the block size of each pixel is set as $5\times5$, and the overlap ratio $Cr$ in the output layer is set as 0.5. For DIFL-FR with two layers of TSK-FS, the number of rules $K_1, K_2$, and the number of outputs $L_1, L_2$, in each TSK-FS layer are optimally set using the search grids $\{2,3,\cdots,10\}$ and $\{4,5,\cdots,16\}$.

*1) ORL Dataset*

The ORL dataset is the most widely used benchmark face dataset. The dataset consists of 400 face images taken by the AT&T Lab from April 1992 to April 1994, with a total of 40 distinct subjects. All the images were taken at different times, with varying lighting, different facial expressions (open closed eyes, smiling / not smiling) and different facial details (glasses / no glasses). For each subject, 10 images were taken in upright and frontal position (with tolerance for side movements). In this section, the size of each image is resized to $32\times32$ pixels, with 256 gray levels. Table III shows the classification results of different methods on the ORL datasets with different proportions of training and test data.

The robustness of DIFL-FR against noise is evaluated with ORL dataset (ORL_Train_8) containing salt & pepper noise and Gaussian noise, see Fig. 2. The results are shown in Table IV, which shows that DIFL-FR is more robust against noise than the other methods.

Next, the robustness of DIFL-FR against occlusion is evaluated with the training set ORL_Train_8 and test sets with an unrelated image randomly located to simulate various levels of contiguous occlusion, from 20% to 60%. See the examples in Fig. 3. It can be seen from the experimental results in Table V that DIFL-FR is superior to the other algorithms under different levels of occlusion. With 20% of pixels occluded, DIFL-FR achieves a high accuracy of 99.75% and still maintains an accuracy of 82.25% even when the occlusion level increases to 60%. It can be concluded that DIFL-FR is robust against noise and occlusion.

*2) Extended Yale B Dataset*

The extended Yale B dataset consists of 2414 images of 38 individuals. For each individual, approximately 64 near frontal images were taken under different illumination conditions. In this study, the size of each image is resized to $32\times32$ pixels, with 256 gray levels. Datasets with different proportions of training and test data are evaluated. The experimental results are shown in Table VI. The classification accuracy of DIFL-FR is found to be superior to that of the other algorithms. For the EYaleB_Train_50 dataset, DIFL-FR even achieves a high accuracy of 99.46%. The strong robustness of DIFL-FR against illumination effect is thus verified by the experiments.

## *B. Large-Scale Datasets*

In this section, the proposed DIFL-FR is evaluated on the large-scale dataset MNIST [25] and Fashion-MNIST [39] datasets.

*1) MNIST Dataset*

The MNIST dataset is a handwritten digital set created by the AT&T Lab. It is composed of 70000 handwritten digits (0 to 9) images, with 60000 training images and 10000 test images. All the digit images were size-normalized to $28\times28$ pixels, with 256 gray levels.

To evaluate the performance of the proposed DIFL-FR, in addition to the methods used previously on small-scale datasets,

TABLE III
COMPARISON OF THE ACCURACY (%) ON ORL DATASETS WITH DIFFERENT PROPORTIONS OF TRAINING AND TEST DATA

| Dataset | Method | | | | | | | | |
|---|---|---|---|---|---|---|---|---|---|
| | DIFL-FR | Raw | BlockHist | LBP | PCA | KPCA | Fisherface | RBM | Autoencoder |
| ORL_Train_2 | **89.44 ±2.35** | 82.5 ±0.96 | 70.44 ±1.93 | 82.31 ±1.95 | 81.75 ±0.81 | 81.56 ±1.96 | 79.75 ±1.58 | 88.5 ±1.5 | 63 ±3.5 |
| ORL_Train_5 | **97.7 ±1.04** | 94.3 ±0.76 | 87.2 ±1.82 | 93.9 ±1.56 | 94.1 ±1.08 | 92.7 ±1.15 | 92.9 ±1.14 | 95.75 ±3.92 | 40 ±6.98 |
| ORL_Train_8 | **99.5 ±0.68** | 98 ±1.9 | 96.75 ±1.9 | 98.25 ±1.43 | 98.25 ±1.68 | 98 ±1.9 | 96.25 ±1.98 | 96.5 ±1.65 | 35 ±8.06 |

TABLE IV
COMPARISON OF THE ACCURACY (%) ON THE ORL_TRAIN_8 DATASET WITH NOISES

| Noise | Method | | | | | | | | |
|---|---|---|---|---|---|---|---|---|---|
| | DIFL-FR | Raw | BlockHist | LBP | PCA | KPCA | Fisherface | RBM | Autoencoder |
| Salt & pepper | **98.5 ±1.37** | 92.75 ±1.63 | 96.75 ±2.44 | 95.75 ±2.59 | 90.25 ±1.05 | 96.75 ±1.43 | 86.25 ±2.34 | 94.5 ±3.57 | 30.25 ±7.31 |
| Gaussian noise with 10 std* | **99.75 ±0.56** | 98.25 ±1.68 | 88.25 ±4.29 | 81 ±2.24 | 98 ±1.9 | 98 ±1.9 | 95.25 ±1.05 | 97.75 ±1.56 | 35.5 ±6.38 |
| Gaussian noise with 30 std | **99 ±3.85** | 94.75 ±2.4 | 56.75 ±2.59 | 19.75 ±4.45 | 94.25 ±2.44 | 96.75 ±1.9 | 88 ±2.27 | 94 ±2.64 | 26 ±3.16 |
| Gaussian noise with 50 std | **93.75 ±3.85** | 83 ±3.6 | 51 ±5.69 | 5.5 ±4.73 | 79.75 ±4.09 | 91 ±3.35 | 70.75 ±3.6 | 88.5 ±3.42 | 15.5 ±8.1 |

* Addition of Gaussian noise with 10, 30 and 50 standard deviations (std), respectively, to the training and test data of the ORL_Train_8 dataset.

TABLE V
COMPARISON OF ACCURACY (%) ON THE ORL_TRAIN_8 DATASET WITH VARYING LEVELS OF OCCLUSION IN THE TEST SET

| Noise | Method | | | | | | | | |
|---|---|---|---|---|---|---|---|---|---|
| | DIFL-FR | Raw | BlockHist | LBP | PCA | KPCA | Fisherface | RBM | Autoencoder |
| 20% Occlusion | **99.75 ±0.56** | 91.25 ±1.77 | 95 ±3.42 | 97 ±2.59 | 90.5 ±2.09 | 94 ±1.05 | 81.75 ±2.88 | 85 ±5.38 | 16.5 ±3.84 |
| 40% Occlusion | **98.25 ±2.09** | 55.75 ±4.56 | 86.5 ±4.37 | 80.5 ±1.12 | 53 ±5.2 | 65 ±4.92 | 41.75 ±2.09 | 68.5 ±4.21 | 13.25 ±2.63 |
| 60% Occlusion | **82.25 ±6.27** | 14.25 ±3.14 | 74.75 ±4.63 | 34.75 ±1.63 | 15.25 ±3.89 | 29.5 ±2.88 | 13 ±3.81 | 57.25 ±6.7 | 12.5 ±3.12 |

TABLE VI
COMPARISON OF ACCURACY (%) ON THE EYALEB DATASET WITH DIFFERENT PROPORTIONS OF TRAINING AND TEST DATA

| Method | Method | | | | | | | | |
|---|---|---|---|---|---|---|---|---|---|
| | DIFL-FR | Raw | BlockHist | LBP | PCA | KPCA | Fisherface | RBM | Autoencoder |
| EYaleB_Train_10 | **84.27 ±1.41** | 81.97 ±2.37 | 43.9 ±0.81 | 80.23 ±0.89 | 81.97 ±2.39 | 71.73 ±1.59 | 83.59 ±1.32 | 88.77 ±1.61 | 10.77 ±2.5 |
| EYaleB_Train_30 | **99.12 ±0.22** | 95.59 ±0.61 | 71.59 ±0.91 | 95.86 ±0.39 | 95.54 ±0.64 | 90.63 ±1.12 | 81.27 ±1.23 | 88.44 ±3.44 | 11.15 ±3.63 |
| EYaleB_Train_50 | **99.46 ±0.16** | 98.44 ±0.36 | 79.53 ±0.87 | 97.82 ±0.16 | 98.48 ±0.52 | 94.59 ±0.1 | 97.59 ±0.56 | 88.43 ±5.22 | 12.06 ±5.15 |

deep learning methods are also included for comparison. The methods include two supervised and three unsupervised deep learning methods. The first supervised deep learning method is a simple CNN (6-2-16-2) structure consisting of two convolutional layers, two pooling layers and two fully connected layers. The two convolutional layers have 6 and 12 convolutional kernels of size $5 \times 5$, respectively. The mean square error is used as the loss function, and the sigmoid function is used as the activation function. The SGD algorithm is used to optimize the network, and the number of training epochs is set as 100. The second supervised deep learning method is the classical CNN structure LeNet-5. The three unsupervised deep learning methods are StrongNet [41], DBM [40], and Deepbit [45]. StrongNet is backpropagation-free architecture with three layers, and its tail layer is trained using a simple linear classifier. The DBM is a Boltzmann machine with multiple hidden layers. DeepBit [45] is a deep learning approach to compact binary descriptor for efficient visual object matching. Table VII shows the classification accuracy of these methods on the MNIST dataset and the proposed DIFL-FR performs the best.

We also compare with the performance of the methods given in the MNIST homepage (http://yann.lecun.com/exdb/mnist/), with attention to 14 two/three-layer models since we consider DIFL-FR with two-layer image feature learning and one output layer. The classification error of the proposed DIFL-FR is 0.61%, outperforming the 14 neural network models whose errors are in the range from 0.7% to 4.7%.

*2) Fashion-MNIST Dataset*

The Fashion-MNIST image dataset is an alternative image dataset of the MNIST dataset. The dataset consists of 70,000 images of different items in 10 categories (T-shirts, trousers, pullovers, skirts, sneakers, etc.). The size, format, and training

TABLE VII
COMPARISON OF ACCURACY (%) ON THE MNIST DATASET

| Traditional Method | Raw | BlockHist | LBP | PCA | KPCA | Fisherface | RBM | Autoencoder |
|---|---|---|---|---|---|---|---|---|
| Accuracy | 92.9 | 91.07 | 94.68 | 91.79 | — | 85.37 | 98.85 | 95.01 |

| Deep Learning Method | CNN | LeNet-5[25] | DBM [40] | StrongNet[41] | Deepbit | DIFL-FR |
|---|---|---|---|---|---|---|
| Accuracy | 98.8 | 99.05 | 99.05 | 98.9 | 97.00 | **99.39** |

Note: In the experiments, the results of KPCA are not available since the latitude of the construction kernel matrix is too large. The results of Lenet-5, DBM and StrongNet are quoted from the results stated in [25], [40], [41] respectively.

TABLE VIII
COMPARISON OF ACCURACY (%) ON THE FASHION-MNIST DATASET

| Traditional Method | Raw | BlockHist | LBP | PCA | KPCA | Fisherface | RBM | Autoencoder |
|---|---|---|---|---|---|---|---|---|
| Accuracy | 84.1 | 81.18 | 82.19 | 83.93 | — | 79.56 | 83.04 | 79.73 |

| Deep Learning Method | CNN | DNN | VGG 16 | Deepbit | DIFL-FR (2) | DIFL-FR (3) |
|---|---|---|---|---|---|---|
| Accuracy | 89.24 | 88.33 | **93.5** | 87.40 | 91.77 | 88.07 |

Note: In the experiments, the results of KPCA are not available since the latitude of the construction kernel matrix is too large. DIFL-FR (2) refers to DIFL-FR with two layers, and DIFL-FR (3) for DIFL-FR with three layers.

TABLE IX
COMPARISON OF ACCURACY (%) ON THE CIFAR-10 DATASET

| Traditional Method | Raw | BlockHist | LBP | PCA | KPCA | Fisherface | RBM | Autoencoder |
|---|---|---|---|---|---|---|---|---|
| Accuracy | 57.82 | 56.38 | 60.19 | 59.21 | — | 53.91 | 59.7 | 58.32 |

| Deep Learning Method | CNN | LeNet-5 | DBM | StrongNet | Deepbit | DIFL-FR |
|---|---|---|---|---|---|---|
| Accuracy | 68.35 | **70.76** | 67.71 | 64.82 | 64.1 | 68.51 |

Note: In the experiments, the results of KPCA are not available since the memory required to build the kernel matrix is too large.

set/test set partitioning of the Fashion-MNIST dataset are identical to those of the MNIST dataset.

Similarly, in addition to the same aforementioned algorithms for the small-scale datasets, four deep learning methods, i.e., CNN, DNN, VGG 16 and Deepbit are also adopted for comparison. The settings for CNN and Deepbit are the same as that for the MNIST Dataset. For DNN, a multiple perceptron network with three hidden layers was adopted where the structure of the hidden layer is 256-128-100. The mean square error is used as the loss function, the tanh function as the activation function, and the SGD as the optimization algorithm for DNN. VGG 16 is a CNN framework with more layers. In the experiment, the standard VGG 16 structure is adopted, where the cross-entropy loss, ReLU activation function and Adam optimization algorithm are adopted, and the number of epochs is set as 100. Table VIII shows the performance of these methods. The DIFL-FR is superior to most of the traditional feature extraction methods and the two/three-layer neural networks, but slightly inferior to the VGG 16 model. However, DIFL-FR is more transparent than VGG 16 due to the fuzzy logic and rules introduced into the model. Note that DIFL-FR is an unsupervised method while the VGG is a supervised method. When DIFL-FR is modified to the supervised version, e.g., by introducing a supervised optimization function, its performance can be further improved. In the experiments, DIFL-FR with two and three layers of TSK-FSs are both considered. It is found, like other deep learning methods, that increasing depth does not necessarily improve the performance. Thus, the depth should be set properly for a specific task. Further analysis on the effectiveness of feature extraction and the effect of hyperparameters on the performance of DIFL-FR can be found in Part F and Part G of the *Supplementary Materials* section.

*3) CIFAR-10 Dataset*

The CIFAR-10 dataset consists of 60,000 RGB images of 10 categories, including automobile, bird, cat, deer, dog, truck and so on, with 50,000 training images and 10,000 test images. All the images are size-normalized to $32 \times 32$ pixels. The differences between CIFAR-10 and MNIST are as follows. The data in CIFAR-10 are 3-channel RGB images whereas the data in MNIST are single-channel grayscale images. The MINST is a dataset of handwritten numbers whereas the CIFAR-10 contains images of real-world objects with noise and of different sizes. Due to these characteristics, general linear classifiers such as linear SVM perform poorly on CIFAR-10. Instead, we choose a nonlinear classifier, i.e., multiple layer perceptron (MLP), to better demonstrate the performance of different feature extraction models.

In the experiments, we compare the proposed DIFL-FR with the same algorithms used previously on the MNIST dataset. For the MLP classifier, we adopt two fully connected layers with the learning rate set to 0.05, cross entropy as the loss function and SGD as the optimization algorithm. The settings of the feature extraction models (i.e., PCA, Autocoder, CNN, Deepbit, etc.) are the same as those in the previous experiments on the MNIST dataset. The comparison results on CIFAR-10 are shown in Table VI. The performance of DIFL-FR is slightly inferior to the supervised algorithm LeNet-5, but it is best among all the unsupervised algorithms and is better than some supervised algorithms.

### C. Ablation Analysis

The proposed DIFL-FR makes use of both the stacked TSK-FSs and the BlockHist for feature extraction. To study their contributions in the performance of DIFL-FR, we conduct ablation study on the ORL_Train_5 dataset. The performance of three versions of DIFL-FR – Version 1 with BlokHist only, Version 2 with stacked TSK-FSs only, and Version 3 with both stacked TSK-FSs and BlockHist (i.e., the proposed DIFL-FR) – are compared. The same classifier, i.e., linear SVM, is used for classification in the three versions. It can be seen from the results in Table X that both modules can help improve the performance, with the stacked TSK-FSs module appearing to make a greater contribution to the performance improvement.

TABLE X
ABLATION ANALYSIS BASED ON ORL_TRAIN_5 DATASET

| Method | | Accuracy |
|---|---|---|
| Version 1： | BlockHist | 87.2±1.82 |
| Version 2： | Stacked TSK-FSs | 96.9±1.78 |
| Version 3： | Stacked TSK-FSs + BlockHist | 97.7±1.04 |

### D. Discussion

Although the existing feature extraction methods can handle image processing tasks relatively efficiently, most of the popular methods are based on deep learning networks which are limited by the quantity and quality of data available, despite the satisfactory performance achieved. For the unsupervised learning tasks concerned in this paper, the existing methods focus only on regenerating the inputs, ignoring the impact of noise on data representation. Moreover, they are unsupervised deep learning methods that are still restricted by the number of unlabeled data. As shown in Table V, while the accuracy of RBM can reach 85.0%, but under noise conditions, it drops to 57.25% only. Furthermore, as shown in Table IV and Table VII, the accuracy of the autoencoder drops from 95.01% for large datasets to 30.25% for smaller datasets. In contrast, the proposed DIFL-FR is based on uncertainty system which is less sensitive to noise and occlusion in image processing tasks. The improved robustness is evident from the results in Table III and Table VI. In addition, DIFL-FR learns features from data with a layer-by-layer mechanism, which can extract features at higher levels for classification and obtain better performance for large datasets (see Table VII and Table VIII). The effectiveness of DIFL-FR in feature extraction is further verified with the experiments and results given in the *Supplementary Materials* section, where the effect of model parameters on the classification performance is also investigated to provide guidelines on the settings and usage of the model.

In the proposed method, clustering and PCA are used to train TSK-FS for feature learning. However, it is not a simple combination of these two techniques, which is elaborated below:

1) The main motivation of our work is to develop a rule-based image feature extraction method which makes the feature extraction more transparent. To achieve this goal, the classical TSK-FS is used as the base model, with which feature learning ability is equipped by integrating clustering and PCA to generate antecedent parameters and learn consequent parameters, respectively. In practice, this strategy is only a feasible solution for training TSK-FS for feature learning. For example, clustering can be replaced with other partition techniques and the PCA criterion can be replaced with other criteria.
2) When clustering and PCA are used to train TSK-FS for feature learning, it is essential to implement a nonlinear model similar to kernel PCA (KPAC). The original input samples are mapped into a high dimensional space by the antecedents of the fuzzy rules (generated using clustering here), and then PCA is implemented to learn a linear model in this high dimensional space (linear model parameters are just the consequent parameters of all fuzzy rules). Thus, the above procedure is very similar to that of KPCA. For TSK-FS, the obtained mapping of the input samples in the high dimensional space is clearly visible and can be easily understood. Conversely, for a kernel method, such KPCA, the mapping in the high dimensional space is usually unknown and invisible, and the corresponding method is like a black box.
3) Furthermore, by introducing stack structure, the deep rule-based feature extraction is realized by our method. Compared with deep PCA, our method has the nonlinear feature learning ability; compared with deep KPCA, our method is more transparent. In addition, our method can be readily extended to adapt to other application scenarios by replacing the PCA criterion with other criteria for consequent parameter learning of TSK-FS. According to the characteristics of the specific task of feature learning, the proposed method can be further developed by introducing more prior information into the optimization criteria.

## V. CONCLUSIONS

The paper proposes a deep image feature learning method with fuzzy rules, which is based on a stacked feature learning structure. The experimental results show that the proposed DIFL-FR method is superior to the traditional hand-crafted feature extraction methods. Compared with classical deep learning methods, the proposed method can achieve satisfactory performance on both small-scale and large-scale datasets. In particular, DIFL-FR not only has a robust nonlinear feature learning ability against noise but also a transparent and interpretable model structure.

Further research on DIFL-FR will be conducted to study the effect of the depth of the model on image feature learning ability and the investigation of the supervised and semi-supervised versions by introducing different learning strategies. For example, we will develop the supervised or semi-supervised optimization criterion for model training.

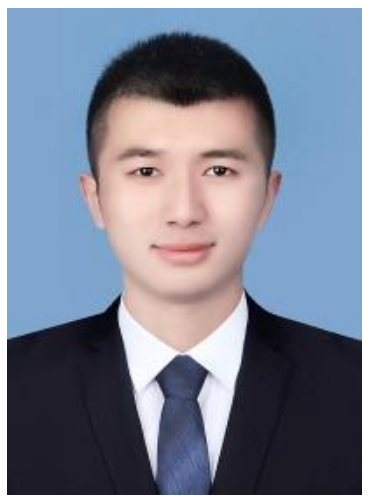

**Xiang Ma** received the B.S. degree in software engineering from Jiangsu Ocean University, Lianyungang, China, in 2017 and the M.S. degree in software engineering from Jiangnan University, Wuxi, China, in 2020. His research interests include computational intelligence, pattern recognition, interpretable artificial intelligence, fuzzy modeling and their applications.

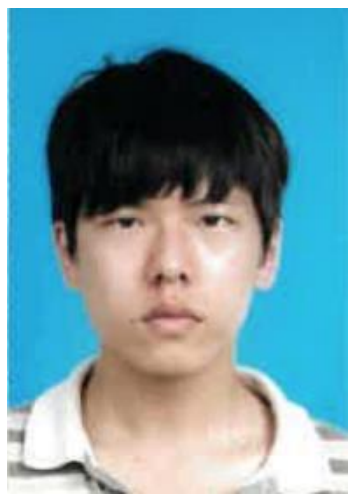

**Liangzhe Chen** received the B.S. degree in Information Security from Northeastern University, Shenyang, China, 2019. He is currently a Ph.D. candidate in the School of Artificial Intelligence and Computer Science, Jiangnan University. His research interests include computational intelligence, machine learning, data augmentation and graph neural networks.

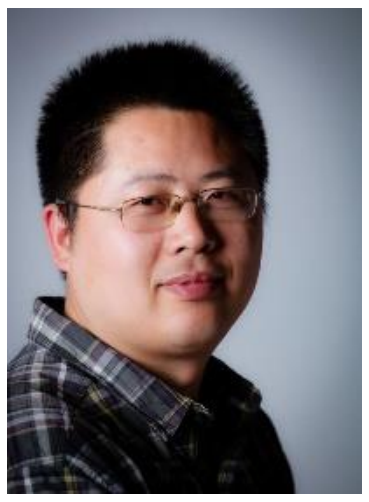

**Zhaohong Deng** (M'2012-SM'2014) received the B.S. degree in physics from Fuyang Normal College, Fuyang, China, in 2002, and the Ph.D. degree in information technology and engineering from Jiangnan University, Wuxi, China, in 2008. He is currently a professor in the School of Artificial Intelligence and Computer Science, Jiangnan University. He has visited University of California-Davis and the Hong Kong Polytechnic University for more than two years. His current research interests include uncertainty modeling, neuro-fuzzy systems, pattern recognition, and their applications. He is the author or coauthor of more than 200 research papers in international/national journals. He has served as an associate editor or guest editor of several international Journals, such as IEEE Transactions on Emerging Topics in Computational Intelligence and Neurocomputing.

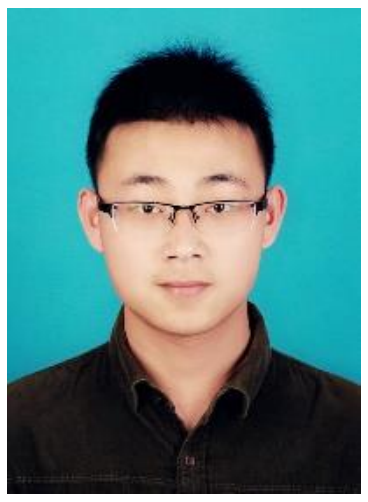

**Peng Xu** received the B.S. degree in digital media and the M.S. degree in software engineering from Jiangnan University, Wuxi, China, in 2017 and 2020, respectively. He is currently an algorithm engineer in Alibaba Inc, Hangzhou, China. His research interests include computational intelligence, machine learning, fuzzy modeling and their applications in dynamic pricing and category planning. He has published several papers in international conference and journals as author and co-author, including AAAI, TFS, TCYB and TNSRE.

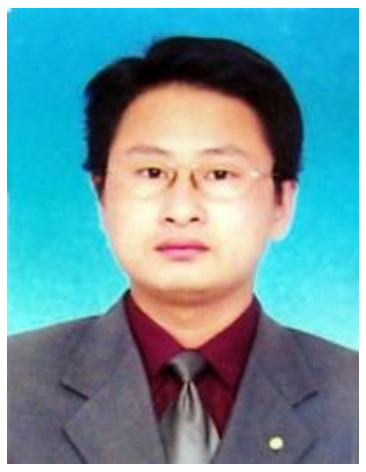

**Qisheng Yan** received the B.S. degree in mathematics from Nanchang University, Nanchang, China, in 1997, and the Ph.D. degree in information technology and engineering from Jiangnan University, Wuxi, China, in 2014. He is currently a Professor in the School of Science, East China University of Technology. His current research interests include neuro-fuzzy systems, machine learning, and their applications. He is the author or coauthor of more than 70 research papers in international/national journals.

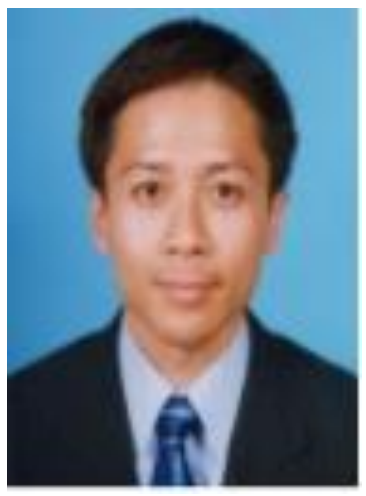

**Kup-Sze Choi** (M'97) received the Ph.D. degree in computer science and engineering from the Chinese University of Hong Kong. He is currently an Associate Professor at the School of Nursing, Hong Kong Polytechnic University and the Director of the Centre for Smart Health. His research interests include virtual reality and artificial intelligence, and their applications in medicine and healthcare.

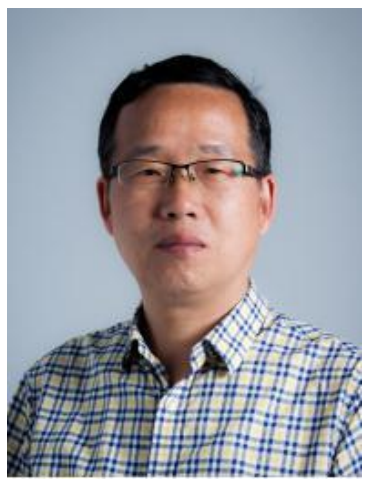

**Shitong Wang** received the M.S. degree in Computer Science from Nanjing University of Aeronautics and Astronautics, China, in 1987. His research interests include artificial intelligence, neuro-fuzzy systems, pattern recognition, and image processing. He has published about 200 papers in international/national journals and has authored seven books.